\documentclass[11pt,a4paper]{article}
\usepackage[hyperref]{naaclhlt2019}
\usepackage{times}
\usepackage{latexsym}

\usepackage{graphicx}
\usepackage[export]{adjustbox}
\usepackage{url}
\usepackage{bm}
\usepackage{here}
\usepackage[fleqn]{amsmath}
\usepackage{amssymb}
\usepackage{amsfonts}
\usepackage{multirow}
\usepackage{arydshln}
\usepackage{enumitem}

\usepackage{algorithm}
\usepackage{algpseudocode}
\usepackage{soul}
\usepackage{float}

\aclfinalcopy 

\title{Accelerated Reinforcement Learning for Sentence Generation\\ by Vocabulary Prediction}

\author{Kazuma Hashimoto\thanks{~~Work was done while the first author was working at the University of Tokyo.} \\
  Salesforce Research \\
  {\small {\tt k.hashimoto@salesforce.com} } \\\And
  Yoshimasa Tsuruoka \\
  The University of Tokyo \\
  {\small {\tt tsuruoka@logos.t.u-tokyo.ac.jp} } \\}

\date{}

\begin{document}
\maketitle

\begin{abstract}
A major obstacle in reinforcement learning-based sentence generation is the large action space whose size is equal to the vocabulary size of the target-side language.
To improve the efficiency of reinforcement learning, we present a novel approach for reducing the action space based on dynamic vocabulary prediction.
Our method first predicts a fixed-size small vocabulary for each input to generate its target sentence.
The input-specific vocabularies are then used at supervised and reinforcement learning steps, and also at test time.
In our experiments on six machine translation and two image captioning datasets, our method achieves faster reinforcement learning ($\sim$2.7x faster) with less GPU memory ($\sim$2.3x less) than the full-vocabulary counterpart.
We also show that our method more effectively receives rewards with fewer iterations of supervised pre-training.

\end{abstract}

\section{Introduction}
Sentence generation with neural networks plays a key role in many language processing tasks, including machine translation~\citep{seq2seq}, image captioning~\citep{mscoco}, and abstractive summarization~\citep{summarization}.
The most common approach for learning the sentence generation models is maximizing the likelihood of the model on the gold-standard target sentences.
Recently, approaches based on reinforcement learning have attracted increasing attention to reduce the gap between training and test situations and to directly incorporate task-specific and more flexible evaluation metrics such as BLEU scores~\citep{bleu} into optimization~\citep{rnn_rl2016}.

While reinforcement learning-based sentence generation is appealing,
it is often too computationally demanding to be used with large
training data.
In reinforcement learning for sentence generation, selecting an action corresponds to selecting a word in the vocabulary $V$.
The number of possible actions at each time step is thus equal to the vocabulary size, which often exceeds tens of thousands.
Among such a large set of possible actions, at most $N$ actions are selected if the length of the generated sentence is $N$, where we can assume $N\ll|V|$.
In other words, most of the possible actions are not selected, and the large action space slows down reinforcement learning and consumes a large amount of GPU memory.

In this paper, we propose to accelerate reinforcement learning by reducing the large action space.
The reduction of action space is achieved by predicting a small vocabulary for each source input.
Our method first constructs the small input-specific vocabulary by selecting $K~(\leq 1000)$ relevant words, and then the small vocabulary is used at both training and test time.

Our experiments on six machine translation and two image captioning datasets show that our method enables faster reinforcement learning with less GPU memory than the standard full softmax method, without degrading the accuracy of the sentence generation tasks.
Our method also works faster at test time, especially on CPUs.
The implementation of our method is available at \url{https://github.com/hassyGo/NLG-RL}.

\section{REINFORCE with Small Vocabularies}

We first describe a neural machine translation model and an image captioning model as examples of sentence generation models.
Machine translation is a text-to-text task, and image captioning is an image-to-text task.
We then review how reinforcement learning is used, and present a simple and efficient method to accelerate the training.

\subsection{Sentence Generation Models}
\label{subsec:baseline}

Recurrent Neural Networks (RNNs) are widely used to generate sentences by outputting words one by one~\citep{seq2seq}.
To generate a sentence $Y=(y_1, y_2, \ldots, y_{N})$, where $N$ is its length, given a source input $X$, a hidden state $h_t\in\mathbb{R}^{d}$ is computed for each time step $t$ $(\geq 1)$ by using its previous information:
\begin{equation}
\label{eq:rnn}
h_t = \mathrm{RNN} \left(h_{t-1}, e(y_{t-1}), s_{t-1} \right),
\end{equation}
where $\mathrm{RNN}(\cdot)$ is an RNN function, $e(y_{t-1})\in\mathbb{R}^{d}$ is a word embedding of $y_{t-1}$, and $s_{t-1}\in\mathbb{R}^{d}$ is a hidden state optionally used to explicitly incorporate the information about the source input $X$ into the transition.
We employ Long Short-Term Memory (LSTM) units~\citep{lstm1} for the RNN function.
The task here is to predict the $t$-th word $y_t$ by computing a target word distribution $p(y|y_{<t}, X)\in\mathbb{R}^{|V|}$, where $|V|$ represents the vocabulary size of the target language.
$p(y|y_{<t}, X)$ is used to generate a sentence by either greedy/beam search or random sampling.

To learn the model parameters, the following cross entropy loss is usually employed:
\begin{equation}
\label{eq:cross_entropy}
L_c(Y_g, X) = -\sum_{t=1}^{N_g}\log p\left(y = y_t|y_{<t}, X \right),
\end{equation}
where we assume that the target sentence $Y_g$ is the gold sequence.
Once we train the model, we can use it to generate unseen sentences.

\paragraph{Machine translation}
In the context of machine translation, the source input $X$ corresponds to a source sentence $(x_1, x_2, \ldots, x_{M})$ of length $M$.
Each word $x_i$ is also associated with a word embedding $\tilde{e}(x_i)\in\mathbb{R}^{d}$.
We assume that a hidden state $\tilde{h}_i\in\mathbb{R}^{2d}$ is computed for each $x_i$ by using a bi-directional RNN with LSTM units~\citep{lstm2}.
That is, $\tilde{h}_i$ is the concatenation of $x_i$'s $d$-dimensional hidden states $[\overrightarrow{h}_i; \overleftarrow{h}_i]$ computed by a pair of forward and backward RNNs.
We set the initial hidden state of the sentence generator as $h_0=\overrightarrow{h}_{M}+\overleftarrow{h}_1$.
Following an attention mechanism proposed in \citet{luong2015}, $s_t$ for predicting $y_t$ is computed as follows:
\begin{eqnarray}
s_t &=& \mathrm{tanh} \left(W_s \left[h_t; \sum_{i=1}^{M} a_i \tilde{h}_i \right] + b_s \right),
\end{eqnarray}
where $a_i=f(h_t, i, \tilde{h})$ is the global-attention function in ~\citet{luong2015}, $W_s\in\mathbb{R}^{d\times 3d}$ is a weight matrix, and $b_s\in\mathbb{R}^{d}$ is a bias vector.
$s_t$ is then used to compute the target word distribution:
\begin{equation}
\label{eq:softmax_mt}
p(y|y_{<t}, X) = \mathrm{softmax} (W_p s_t + b_p),
\end{equation}
where $W_p\in\mathbb{R}^{|V|\times d}$ is a weight matrix, and $b_p\in\mathbb{R}^{|V|}$ is a bias vector.

\paragraph{Image captioning}
In the case of image captioning, the source input $X$ corresponds to an image to be described.
We assume that in our pre-processing step, each input image is fed into a convolutional neural network to extract its fixed-length feature vector $f\in\mathbb{R}^{d_f}$.
More specifically, we use the pre-computed feature vectors provided by \citet{kiros2014}, and the feature vectors are never updated in any model training processes.
The input feature vector is transformed into the initial hidden state $h_0 = \mathrm{tanh}\left( W_f f + b_f \right)$, where $W_f\in\mathbb{R}^{d\times d_f}$ is a weight matrix, and $b_f\in\mathbb{R}^{d}$ is a bias vector.
In contrast to machine translation, we do not use $s_{t-1}$ in Equation~(\ref{eq:rnn}); more concretely, we do not use any attention mechanisms for image captioning.
Therefore, we directly use the hidden state $h_t$ to compute the target word distribution:
\begin{equation}
\label{eq:softmax_cap}
p(y|y_{<t}, X) = \mathrm{softmax} (W_p h_t + b_p),
\end{equation}
where the weight and bias parameters are analogous to the ones in Equation~(\ref{eq:softmax_mt}).

For both of the tasks, we use the weight-tying technique~\citep{tai_softmax,tai_softmax_2} by using $W_p$ as the word embedding matrix.
That is, $e(y_t)$ is the $y_t$-th row vector in $W_p$, and the technique has shown to be effective in machine translation~\citep{lgp_nmt} and text summarization~\cite{romain2018}.

\subsection{Applying Reinforcement Learning}
\label{subsec:rl}

One well-known limitation of using the cross entropy loss in Equation~(\ref{eq:cross_entropy}) is that the sentence generation models work differently at the training and test time.
More concretely, the models only observe gold sequences at the training time, whereas the models have to handle unseen sequences to generate sentences at the test time.

To bridge the gap, reinforcement learning has started gaining much attention~\citep{rnn_rl2016,gnmt,self_critic2017,simp_rl2017,romain2018,gan_rl2018}.
In this work, we focus on the most popular method called REINFORCE~\citep{REINFORCE}.\footnote{We tried self critic~\citep{self_critic2017}, but did not observe significant improvement over REINFORCE.}
In REINFORCE, the sentence generation model sets an initial state given a source input, and then iterates an action selection and its corresponding state transition.
The action selection corresponds to randomly sampling a target word from Equation~(\ref{eq:softmax_mt}) and (\ref{eq:softmax_cap}), and the state transition corresponds to the RNN transition in Equation~(\ref{eq:rnn}).

Once a sentence is generated, an approximated loss function is defined as follows:
\begin{equation}
L_r(Y, X)= -\sum_{t=1}^{N} R_t \log{p(y = y_t | y_{<t}, X)},
\end{equation}
where $R_t$ is the reward at time step $t$, and the loss is approximated by the single example $Y$.
$R_t$ is used to evaluate how good the $t$-th action selection is.
Unlike maximum likelihood training, the reward function can be defined by using task-specific evaluation scores like BLEU for machine translation.
In this paper, we employ GLEU proposed by \citet{gnmt}, a variant of sentence-level BLEU.
Following the implementation in \citet{rnn_rl2016}, we define $R_t = \mathrm{GLEU}(Y, Y_g) - b_t$,
where $b_t$ is a baseline value estimating the future reward from the next time step to reduce the variance of the gradients.
To estimate $b_t$, we jointly train a linear regression model by minimizing $\|b_t - \mathrm{GLEU}(Y, Y_g)\|^2$, and $b_t$ is computed as $b_t = \sigma (W_r \cdot s_t + b_r)$,
where $W_r\in\mathbb{R}^{d}$ is a weight vector, $b_r$ is a bias, $\sigma(\cdot)$ is the logistic sigmoid function, and in the case of image captioning, $h_t$ is used instead of $s_t$.

\paragraph{Overall model training}
The reinforcement learning step is usually applied after pre-training the models with the cross entropy loss in Equation~(\ref{eq:cross_entropy}).
At the REINFORCE phase, we define the following joint loss function:
\begin{eqnarray}
\label{eq:lambda}
L = \lambda L_c + (1-\lambda) L_r,
\end{eqnarray}
where $\lambda$ is a hyperparameter, and $\lambda=0.0$ usually leads to unstable training~\citep{gnmt}.

\subsection{Large Action-Space Reduction}
The vocabulary size $|V|$ is usually more than ten thousands for datasets covering many sentences with a variety of topics.
However, for example, at most 100 unique words are selected when generating a sentence of length 100.
That is, the output length $N$ is much smaller than the vocabulary size $|V|$, and this fact motivated us to reduce the large action space.
Moreover, we have in practice found that REINFORCE runs several times slower than the supervised learning with the cross entropy loss.

To accelerate the training, we propose to construct a small action space for each source input.
In other words, our method selects a small vocabulary $V'$ of size $K$ for each source input in advance to the model training.
In this section, we assume that $V'$ is given and represented with a sparse binary matrix $M_X\in\mathbb{R}^{K\times |V|}$, where there are only $K$ non-zero elements at position $(i, w_i)$ for $1 \leq i \leq K$.
$w_i$ is a unique word index in $V$.
$M_X$ is used to construct a small subset of the parameters in the softmax layer:
\begin{eqnarray}
W_p' = M_X W_p,~~~b_p' = M_X b_p, 
\end{eqnarray}
and $W_p'\in\mathbb{R}^{K\times d}$ and $b_p'\in\mathbb{R}^{K}$ are used instead of $W_p$ and $b_p$ in Equation~(\ref{eq:softmax_mt}) and (\ref{eq:softmax_cap}).
Therefore, in mini-batched processes with a mini-batch size $B$, our method constructs $B$ different sets of $(W_p', b_p')$.

\paragraph{Relationship to previous work}
Sampling-based approximation methods have previously been studied to reduce the computational cost at the large softmax layer in probabilistic language modeling~\citep{blackout,nce_gpu}, and such methods are also used to enable one to train neural machine translation models on CPUs~\cite{eriguchi2016}.
The construction of $(W_p', b_p')$ in our method is similar to these softmax approximation methods in that they also sample small vocabularies either at the word level~\citep{blackout}, sentence level~\citep{lgp_nmt}, or mini-batch level~\cite{nce_gpu}.
However, one significant difference is that the approximation methods work only at training time using the cross entropy loss, and full softmax computations are still required at test time.
The difference is crucial because a sentence generation model needs to simulate its test-time behavior in reinforcement learning.

\section{Target Vocabulary Prediction}

The remaining question is how to construct the input-specific vocabulary $V'$ for each source input $X$.
This section describes our method to construct $V'$ by using a vocabulary prediction model which is separated from the sentence generation models.

\subsection{Input Representations}

In the vocabulary prediction task, the input is the source $X$ (source sentences or images) to be described, and the output is $V'$.
We should be careful {\it not} to make the prediction model computationally expensive; otherwise the computational efficiency by our method would be canceled out.

To feed the information about $X$ into our vocabulary prediction model, we define an input vector $v(X)\in\mathbb{R}^{d_v}$.
For image captioning, we use the feature vector $f$ described in Section~\ref{subsec:baseline}: $v(X) = W_v f + b_v$,
where $W_v\in\mathbb{R}^{d_v\times d_f}$ is a weight matrix, and $b_v\in\mathbb{R}^{d_v}$ is a bias vector.
For machine translation, we employ a bag-of-embeddings representation:
$v(X) = \frac{1}{M}\sum_{i=1}^{M} \tilde{e}_v(x_i)$,
where the $d_v$-dimensional word embedding $\tilde{e}_v(x_i)\in\mathbb{R}^{d_v}$ is different from $\tilde{e}(x_i)$ used in the machine translation model.
By using the different set of the model parameters, we avoid the situation that our vocabulary prediction model is affected during training the sentence generation models.

\paragraph{Relationship to previous work}
Vocabulary prediction has gained attention for training sequence-to-sequence models with the cross entropy loss~\citep{word_pred_emnlp,word_pred_aaai}, but not for reinforcement learning.
Compared to our method, previous methods jointly train a vocabulary predictor by directly using source encoders as input to the predictor.
One may expect joint learning to improve both of the vocabulary predictor and the sentence generator, but in practice such positive effects are not clearly observed.
\citet{word_pred_emnlp} reported that the joint learning improves the accuracy of their machine translation models, but our preliminary experiments did not indicate such accuracy gain.
Such a joint training approach requires the model to continuously update the vocabulary predictor during REINFORCE, because the encoder is shared.
That is, the action space for each input changes during reinforcement learning, and we observed unstable training.
Therefore, this work separately models the vocabulary predictor and focuses on the effects of using the small vocabularies for REINFORCE.

Another note is that \citet{large_vocab} and \citet{word_pred_arxiv} also proposed to construct small vocabularies in advance to the cross entropy-based training.
They suggest that the use of word alignment works well, but using the word alignment is not general enough, considering that there exist different types of source input.
By contrast, our method can be straightforwardly applied to the two sentence generation tasks with the different input modalities (i.e. image and text).

\subsection{Multi-Label Classification}

Once the input representation $v(X)$ is computed, we further transform it by a single residual block~\citep{resnet}: $r(X) = \mathrm{Res}\left(v(X) \right)\in\mathbb{R}^{d_v}$.\footnote{We can use arbitrary types of hidden layers or even linear models like SVMs, but we found this one performed the best. We describe the details of this in the supplemental material.}
Then $r(X)$ is fed into a prediction layer:
\begin{eqnarray}
\label{eq:voc_pred}
o = \sigma \left(W_o r(X) + b_o \right),
\end{eqnarray}
where $W_o\in\mathbb{R}^{|V|\times d_v}$ is a weight matrix, and $b_o\in\mathbb{R}^{|V|}$ is a bias vector.
The $i$-th element $o_i$ corresponds to the probability that the $i$-th word in the target vocabulary $V$ appears in the target sentence $Y$ given its source $X$.

We use the training data for the sentence generations tasks to train the vocabulary predictor.
For each $X$ in the training data, we have its gold target sentence $Y_g$.
We train the vocabulary predictor as a multi-label classification model by the following loss function:
\begin{equation}
-\sum_{i=1}^{|V|} \left( t_i \log o_i + (1-t_i) \log (1-o_i) \right),
\end{equation}
where $t_i$ is equal to $1.0$ if the $i$-th word in $V$ is included in $Y_g$, and otherwise $t_i$ is $0.0$. 
In practice, we apply the label smoothing technique~\citep{smoothing} to the loss function.

We evaluate the accuracy of the vocabulary predictor by using a separate development split $D$:
\begin{equation}
\frac{\#~\mathrm{of~correctly~predicted~words~in}~D}{\#~\mathrm{of~words~in}~D},
\end{equation}
where we select the top-$K$ predictions in Equation~(\ref{eq:voc_pred}) for each source input $X$ in $D$, and the evaluation metric is a recall score.
We use the top-$K$ words to construct the input-specific vocabularies $V'$ for the sentence generation models, and we restrict that the recall is $100\%$ for the training data.

\section{Experimental Settings}

We describe our experimental settings, and the details can be found in the supplemental material.

\subsection{Datasets}
We used machine translation datasets of four different language pairs: English-to-German (En-De), English-to-Japanese (En-Ja), English-to-Vietnamese (En-Vi), and Chinese-to-Japanese (Ch-Ja).
For image captioning, we used two datasets: MS COCO~\citep{mscoco} and Flickr8K.
Table~\ref{tb:dataset} summarizes the statistics of the training datasets, where the number of training examples (``Size''), the target vocabulary size ($|V|$), and the maximum length of the target sentences ($\max(N)$) are shown.
For the machine translation datasets, we manually set $\max(N)$ and omitted training examples which violate the constraints.

\begin{table}[t]
  \begin{center}
{\small
    \begin{tabular}{l|ccc}
	Dataset & Size & $|V|$ & $\max(N)$ \\ \hline
	En-De & ~~~100,000 & 24,482 & ~~50 \\
	En-Ja (100K) & ~~~100,000 & 23,536 & ~~50 \\
	En-Ja (2M) & 1,998,821 & 70,668 & 100 \\
	En-Ja (2M, SW) & 1,998,816 & 37,905 & 200 \\
	En-Vi & ~~~132,406 & 14,321 & 100 \\
	Ch-Ja & ~~~100,000 & 23,383 & ~~50 \\ \hdashline
	MS COCO & ~~~413,915 & 14,543 & ~~57 \\
	Flickr8K & ~~~~~30,000 & ~~4,521 & ~~38 \\ \hline
    \end{tabular}
}
    \caption{Statistics of the training datasets.}
    \label{tb:dataset}
  \end{center}
\end{table}

\begin{table*}[t]
  \begin{center}
    \begin{tabular}{ll|cc|cc}
				& & \multicolumn{2}{c|}{Cross entropy} & \multicolumn{2}{c}{REINFORCE w/ cross entropy} \\
				& & Small softmax & Full softmax & Small softmax & Full softmax \\ \hline
	\multirow{4}{*}{Translation}
	& En-De & 11.09$\pm$0.51 & 10.84$\pm$0.37 & 12.13$\pm$0.33 & 11.73$\pm$0.23 \\
	& En-Ja (100K) & 28.26$\pm$0.15 & 28.05$\pm$0.40 & 29.14$\pm$0.13 &29.01$\pm$0.35   \\
	& En-Vi & 24.56$\pm$0.14 & 24.53$\pm$0.18 & 24.98$\pm$0.11 & 24.92$\pm$0.09   \\
	& Ch-Ja & 29.27$\pm$0.08 & 28.97$\pm$0.15 & 30.10$\pm$0.12 &29.80$\pm$0.15   \\ \hdashline
	\multirow{2}{*}{Image captioning}
	& MS COCO & 24.88$\pm$0.25 & 24.75$\pm$0.36 & 26.43$\pm$0.32 & 25.74$\pm$0.13 \\
	& Flickr8K   & 16.45$\pm$0.28 & 16.52$\pm$0.11 & 19.04$\pm$0.43 & 19.17$\pm$0.24 \\ \hline

    \end{tabular}
    \caption{BLEU scores for the development splits of the six datasets.
			``Small softmax'' corresponds to our method.}
    \label{tb:bleu_small}
  \end{center}
\end{table*}

\noindent
{\bf En-De:}
We used 100,000 training sentence pairs from {\tt news commentary} and {\tt newstest2015} as our development set, following \citet{eriguchi2017}.

\noindent
{\bf En-Ja:}
We used parallel sentences in ASPEC~\citep{aspec} and constructed three types of datasets: En-Ja (100K), En-Ja (2M), and En-Ja (2M, SW).
The 100K and 2M datasets were constructed with the first 100,000 and 2,000,000 sentence pairs, respectively.
To test our method using subword units, we further pre-processed the 2M dataset by using the SentencePiece toolkit~\citep{spm} to construct the En-Ja (2M, SW) dataset.

\noindent
{\bf En-Vi:}
We used the pre-processed datasets provided by \citet{envi}. 
Our development dataset is the {\tt tst2012} dataset.

\noindent
{\bf Ch-Ja:}
We constructed the Ch-Ja dataset by using the first 100,000 sentences from {\tt ASPEC}.

\noindent
{\bf MS COCO and Flickr8K:}
We used the pre-processed datasets provided by \citet{kiros2014}. 
We can also download the 4096-dimensional feature vectors $f$ (i.e., $d_f = 4096$).

\subsection{Settings of Vocabulary Prediction}
We set $d_v=512$ for all the experiments.
We used {\tt AdaGrad}~\cite{duchi2011} to train the vocabulary predictor with a learning rate of $0.08$ and a mini-batch size of 128.
The model for each setting was tuned based on recall scores (with $K=1000$) for the development split.

\subsection{Settings of Sentence Generation}
We set $d=256$ with single-layer LSTMs for all the experiments, except for the En-Ja (2M) and (2M, SW) datasets.
For the larger En-Ja datasets, we set $d=512$ with two-layer LSTMs.
We used stochastic gradient decent with momentum, with a learning rate of $1.0$, a momentum rate of $0.75$, and a mini-batch size of 128.
The model for each setting was tuned based on BLEU scores for the development split.
All of the models achieved the best BLEU scores for all the datasets within 15 to 20 training epochs.
Each of the selected models with the best BLEU scores was used for the following REINFORCE step.
For REINFORCE, we set $\lambda=0.005$, and the learning rate was set to $0.01$.
The REINFORCE steps required around 5 epochs to significantly improve the BLEU scores.

\subsection{Computational Resources and Mini-Batch Processing}
\label{subsec:gpu}
We used a single GPU of {\tt NVIDIA GeForce GTX 1080}\footnote{The GPU memory capacity is 11,178MiB.} to run experiments for the En-De, En-Ja (100K), En-Vi, Ch-Ja, MS COCO, and Flickr8K datasets.
For the En-Ja (2M) and En-Ja (2M, SW) datasets, we used a single GPU of {\tt NVIDIA Tesla V100}\footnote{The GPU memory capacity is 16,152MiB (AWS p3).} to speedup our experiments.

\paragraph{Mini-batch splitting}
It should be noted that our small softmax method can be run even on the single {\tt GTX 1080} GPU for the larger translation datasets, whereas the full softmax method runs out of the GPU memory.
A typical strategy to address such out-of-memory issues is to use multiple GPUs, but we have found that we need at most eight GPUs to conduct our experiments on the full softmax method with REINFORCE.\footnote{This also depends on the mini-batch size.}
Moreover, using the multiple GPUs does not always speedup the training time.
We instead employ another strategy to split the mini-batch at each training iteration.
First, we sort the mini-batch examples according to the lengths of the source (or target) text, and then split the mini-batch into $S$ sets of the training examples.
For example, in our case the mini-batch size is 128, and if $S$ is set to 4, each of the smaller sets includes 32 training examples.
We perform back-propagation for each set one by one, and at each step we delete the corresponding computational graphs to reduce the GPU memory consumption.
Finally, the accumulated partial derivatives are used to update the model parameters.
More details can be found in our Pytorch 0.4 implementation.

\section{Results of Sentence Generation Tasks}
\label{sec:res_voc}

\begin{figure}[t]
	\begin{center}
    	\includegraphics[width=75mm]{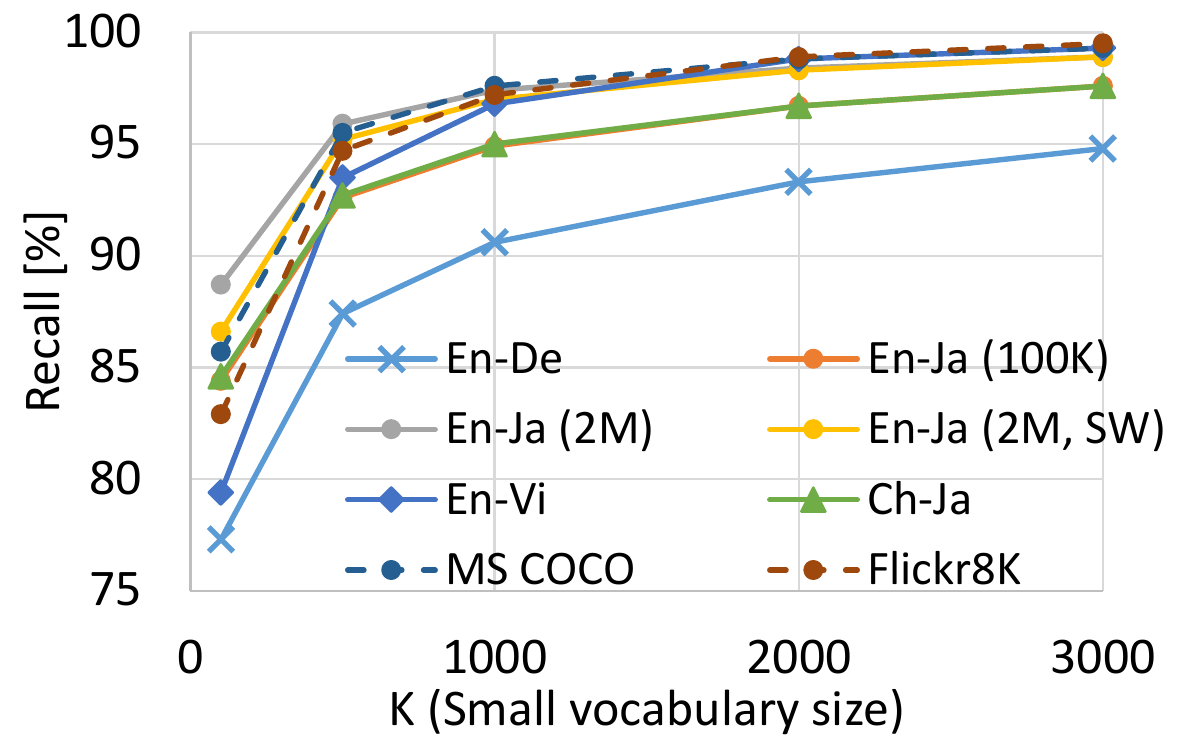}
    \end{center}
\caption{Recall scores of our vocabulary predictor.}
\label{fig:voc_pred}
\end{figure}

\subsection{Accuracy of Vocabulary Prediction}

Figure~\ref{fig:voc_pred} shows recall scores with respect to different values of the small vocabulary size $K$ for each dataset.
We can see that the recall scores reach 95\% with $K=1000$ for most of the datasets.
One exception is the En-De dataset, and this is not surprising because a German vocabulary would become sparse by many compound nouns.

These results show that our vocabulary predictor works well for source inputs of different modalities (text and image) and their corresponding different target languages.
Our method also works at the subword level as well as at the standard word level.
For training the sentence generation models, we set $K=500$ for the Flickr8K dataset and $K=1000$ for the other datasets.
Our empirical recommendation is $K=1000$ if $|V|$ is larger than 10,000 and otherwise $K=500$.

\subsection{Accuracy of Sentence Generation}

The goal of this paper is achieving efficient reinforcement learning for sentence generation to encourage future research, but before evaluating the efficiency of our method, we show that using the small vocabularies does not degrade the accuracy of the sentence generation models.
Table~\ref{tb:bleu_small} shows BLEU scores for the development splits of the four machine translation and two image captioning datasets.
The BLEU scores are averaged over five different runs with different random seeds, and the standard deviations are also reported.

We can see in Table~\ref{tb:bleu_small} that our method (Small softmax) keeps the BLEU scores as high as those of ``Full softmax''.
For some datasets, the BLEU scores of our method are even better than those of the full softmax method.
The trend is consistent in both of the cross entropy training phase and the REINFORCE phase.
These results indicate that our method works well for different machine translation and image captioning datasets.
We also confirmed that our experimental results are competitive with previously reported results when using the same training datasets; for example, our En-Vi test set result on {\tt tst2013} is 27.87$\pm$0.21 (cf. 26.9 in \citet{envi}).

\begin{figure}[t]
	\begin{center}
    	\includegraphics[width=70mm]{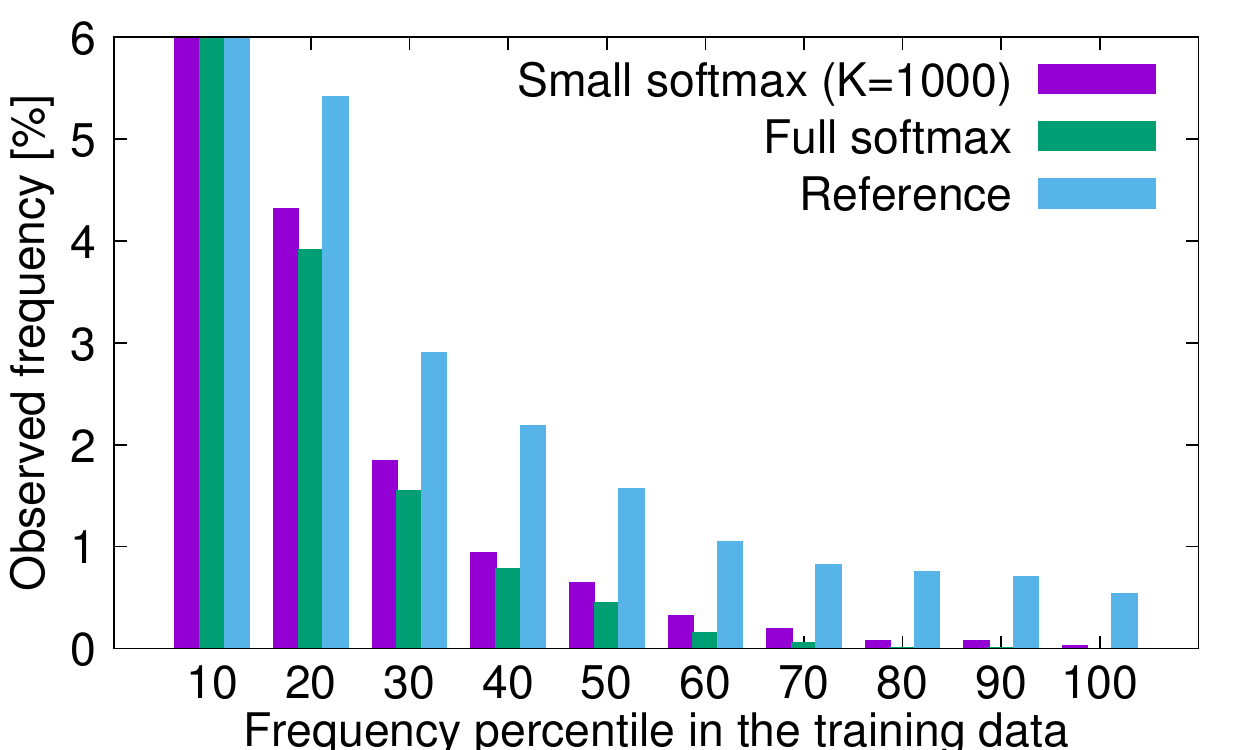}
    \end{center}
\caption{An analysis on the En-De translation results.
The small and full softmax results are based on the REINFORCE setting, and ``Reference'' corresponds to the En-De reference translations of the development split.}
\label{fig:histgram}
\end{figure}

\paragraph{Better generation of rare words}
These BLEU scores suggest that our method for reinforcement learning has the potential to outperform the full softmax baseline.
However, it is still unclear what is the potential advantage in terms of generation quality.
We therefore analyzed the differences between output sentences of the small and full softmax methods, following \citet{uncertainty}.
Figure~\ref{fig:histgram} shows the results of the En-De translation dataset, and we observed the same trend for all the other datasets.
Each entry is computed as follows:
\begin{equation}
\frac{\#~\mathrm{of~output~words~in~each~percentile}}{\#~\mathrm{of~output~words}},
\end{equation}
where the ``10'' percentile includes the top 10\% of the most {\it frequent} words, and the ``100'' percentile includes the top 10\% of the most {\it infrequent} words.
We can see that our small softmax method better outputs rare words, and these results suggest that using input-specific vocabularies is useful in controlling action spaces for reinforcement learning.

\begin{figure}[t]
	\begin{center}
    	\includegraphics[width=70mm]{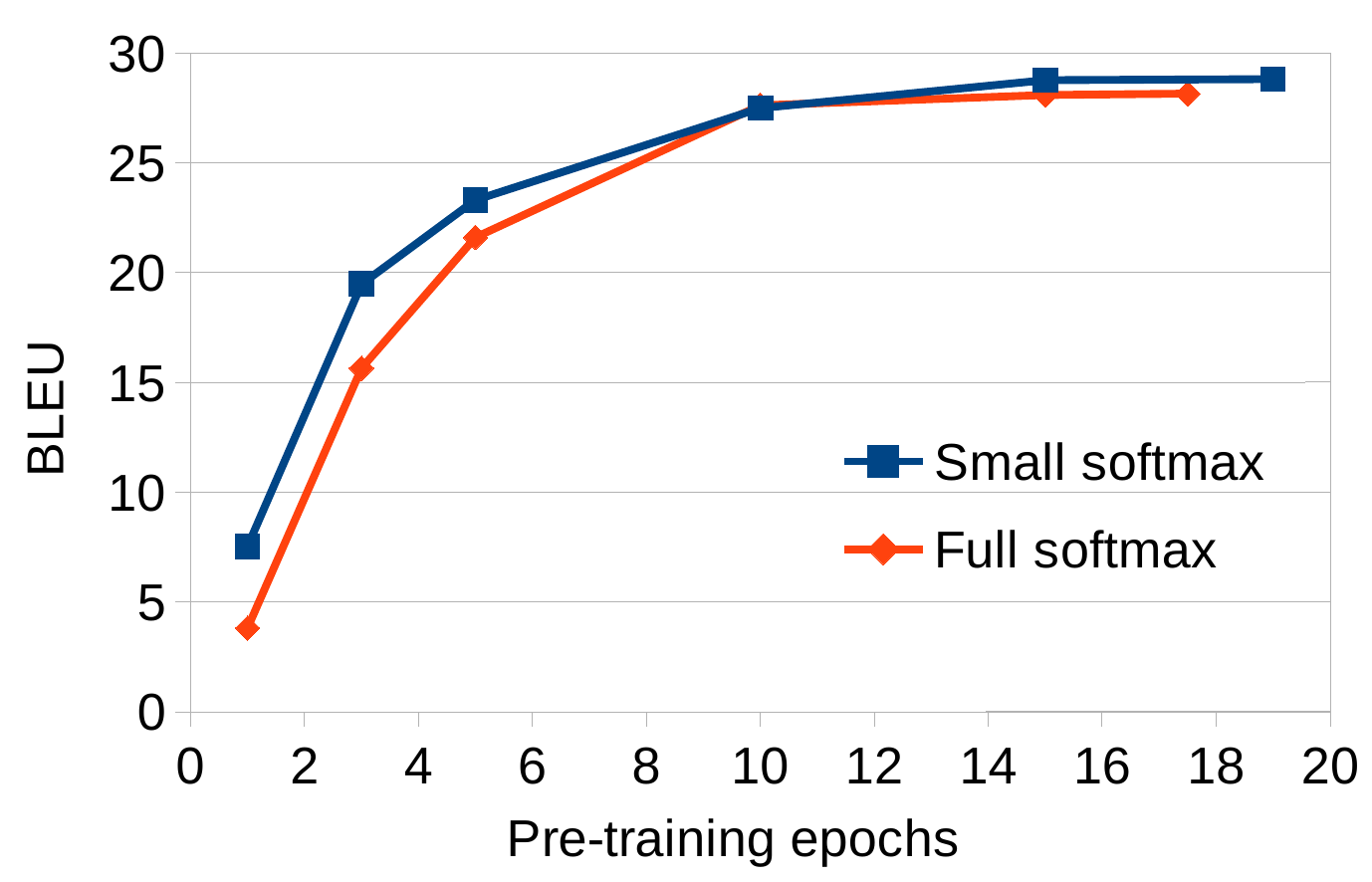}
    \end{center}
\caption{Effects of pre-training for REINFORCE.}
\label{fig:reinforce}
\end{figure}

\paragraph{Effectiveness with fewer pre-training steps}
We followed the standard practice that the models are pre-trained by maximum likelihood before starting reinforcement learning.
However, such pre-training may have a negative effect in reinforcement learning.
Consider the situation where the pre-training leads to zero cross-entropy loss.
In this case, nothing will be learned during reinforcement learning because no exploratory action can be performed.
Although pre-training in practice does not lead to zero cross-entropy loss, it can still overfit the data and result in very sharp output distributions, thereby hindering exploration
in reinforcement learning.
It is therefore important to consider a reinforcement learning setting with less or no pre-training~\citep{web}.
In Figure~\ref{fig:reinforce} for the En-Ja (100K) dataset, we show that the small softmax method works more effectively with fewer pre-training epochs.
For this experiment, we set $\lambda=0$ in Equation~(\ref{eq:lambda}) to purely focus on REINFORCE.
Using GLEU (or BLEU) scores gives sparse rewards, and thus the resulting BLEU scores are very low with fewer pre-training steps, but the small softmax method has the potential to work well if we can design more effective reward functions.

\paragraph{Results on larger datasets}

\begin{table}[t]
  \begin{center}
{\small
    \begin{tabular}{l|cc}
						& 2M & 2M, SW  \\ \hline
	Cross entropy		& 38.76  & 39.15 \\
	w/ beam search 	& 39.88 & 40.35 \\ \hdashline
	REINFORCE w/ cross entropy	     & 40.10 & 40.26   \\
	w/ beam search   & 40.36 & 40.38 \\
	w/ beam search ($K$=~~500)& 40.07 & 40.07 \\
	w/ beam search ($K$=2000) & 40.30 & 40.50 \\
	w/ beam search ($K$=3000) & 40.27 & 40.41 \\ \hline
    \end{tabular}
}
    \caption{BLEU scores for the development split of the En-Ja (2M) and En-Ja (2M, SW) datasets.}
    \label{tb:bleu_large}
  \end{center}
  \begin{center}
{\small
    \begin{tabular}{l|c}
						&  \\ \hline
	REINFORCE w/ cross entropy ($K$=1000)  & 40.16 \\ 
	w/ beam search 	& 40.50 \\ \hdashline 
	- Cross entropy (1.3M) w/ beam search		 & \multirow{2}{*}{39.42}  \\
	\citep{lgp_nmt}  \\
	- Cross entropy (2M) w/ beam search   	& \multirow{2}{*}{40.29}  \\
	\citep{wat_oda}  \\
	- Cross entropy (2M+1M back-trans.)  & \multirow{2}{*}{41.42} \\
	w/ beam search~\citep{wat_ntt}  \\ \hline
    \end{tabular}
}
    \caption{BLEU scores for the En-Ja test split, where we use the En-Ja (2M, SW) dataset.
         The 95\% confidence interval by bootstrap resampling~\citep{bootstrap} is [39.61, 41.47] for our beam search result.}
    \label{tb:bleu_test}
  \end{center}
\end{table}

\begin{table*}[t]
  \begin{center}
{\small
    \begin{tabular}{l|cc|cc|cc|cc|cc}
                 & & & \multicolumn{4}{c|}{Training time [minutes/epoch]} & \multicolumn{4}{c}{GPU memory [MiB]} \\ \hdashline
				& &  & \multicolumn{2}{c|}{CE} & \multicolumn{2}{c|}{REINFORCE w/ CE} & \multicolumn{2}{c|}{CE} & \multicolumn{2}{c}{REINFORCE w/ CE} \\
				& $|V|$ & $\max{(N)}$ & Small & Full & Small & Full & Small & Full & Small & Full \\ \hline
	En-Ja (100K) & 23,536 & ~~50 &  ~~~4.6 & ~~~4.8 & ~~10.1 & ~~21.2 & 1,781 & ~~6,061 & 2,193 & ~~7,443   \\
	En-Ja (2M) & 70,668 & 100 &  ~~95.7 & 141.4 & 231.3 & 635.9 & 5,033 & 10,527 & 6,485 & 14,803   \\
	En-Vi & 14,321 & 100 &  ~10.5 & ~~10.7 & ~~23.2 & ~~38.4  &  2,149 & 10,645 & 2,909 & 10,807   \\\hdashline
	MS COCO & 14,543 & ~~57 & ~~~~4.4 & ~~~4.2 & ~~11.6 & ~~22.9  &   1,419 & ~~8,587 & 1,785 & 10,651 \\
	Flickr8K   & ~~4,521 & ~~38 & ~~~~0.3 & ~~~0.3 & ~~~~0.8 & ~~~~0.9   &   ~~~911 & ~~2,031 & 1,031 & ~~3,197  \\ \hline

    \end{tabular}
}
    \caption{Training time, and maximum memory consumption on our GPU devices for the text generation models.
    For the full softmax baseline on the En-Ja (2M) experiments, the mini-batch splitting strategy (described in Section~\ref{subsec:gpu}) is applied.
    CE: Cross-Entropy, Small: Small softmax (our proposed method), Full: Full softmax (the baseline).}
    \label{tb:time}
  \end{center}
\end{table*}

To see whether our method works in larger scales, Table~\ref{tb:bleu_large} shows BLEU scores for the development split when using the En-Ja (2M) and En-Ja (2M, SW) datasets.\footnote{For the 2M dataset, the full softmax baseline achieves BLEU scores of 38.67 and 39.84 for the ``Cross entropy'' and ``REINFORCE w/ cross entropy'' settings, respectively.}
These results show that our method consistently works even on these larger datasets at the word and subword levels.
In this table we also report how our method works with beam search,
and the greedy-based BLEU scores are very close to those of beam search after the REINFORCE phase.
When performing a beam search, we can optionally use different sizes of the small vocabulary, but we observe that our method is robust to the changes, whereas \citet{word_pred_aaai} reported that their dynamic vocabulary selection method is sensitive to such changes.

For reference, we report the test set results in Table~\ref{tb:bleu_test}.
We cite BLEU scores from previously published papers which reported results of single models (i.e., without ensemble).
Our method with greedy translation achieves a competitive score.
It should be noted that \citet{wat_ntt} achieve a better score presumably because they used additional in-domain one million parallel sentences obtained by the back-translation technique~\citep{backtrans}.

\section{Efficiency of the Proposed Method}

This section discusses our main contribution: how efficient our method is in accelerating reinforcement learning for sentence generation.

\subsection{Speedup at Training Time}

We have examined the training-time efficiency of our method.
Table~\ref{tb:time} shows the training time [minutes/epoch] for five different datasets.
We selected the five datasets to show results with different vocabulary sizes and different maximum sentence lengths, and we observed the same trend on the other datasets.
The vocabulary size $|V|$ and the maximum sentence length $\max(N)$ are shown for each training dataset.
In the training with the standard cross entropy loss, the speedup by our method is not impressive as long as the vocabulary size $|V|$ can be easily handled by the GPUs.
We set $S=2$ for the cross entropy training of the ``Full softmax'' method in the En-Ja (2M) setting, to reduce the GPU memory consumption as described in Section~\ref{subsec:gpu}.

In the training with the REINFORCE algorithm, the speedup by our method is enhanced.
In particular, in the En-Ja (2M) experiments, our method gains a factor of 2.7 speedup compared with the full softmax baseline ($S=3$).
For most of the experimental settings, the speedup significantly accelerates our research and development cycles when working on reinforcement learning for sentence generation tasks.
One exception is the Flickr8K dataset whose original vocabulary size $|V|$ is already very small, and the lengths of the target sentences are short.
In the supplementary material, we also show the test-time efficiency.

\subsection{GPU Memory Consumption}
\label{subsec:memory}

Our method is also efficient in terms of GPU memory consumption at training time.
Table~\ref{tb:time} also shows the maximum GPU memory consumption during the training.
These results show that our method easily fits in the memory of the single {\tt GTX 1080} GPU, whereas ``Full softmax'' is very sensitive to the vocabulary size $|V|$ and the sentence lengths.
In particular, we observe about 56\% reduction in memory usage when using the En-Ja (2M) dataset.
By saving the memory usage, one could try using larger models, larger mini-batches, larger vocabularies, and longer target sentences without relying on multiple GPUs.

\paragraph{Scalability of our method}
To further show the memory efficiency our our method, we measured the GPU memory consumption with a larger mini-batch size, 2048.
We applied the mini-batch splitting strategy to both the small and full softmax methods to handle such a large mini-batch size.
In the En-Ja (2M) experiments with REINFORCE, our small softmax method works with the large batch-size by setting $S=6$, whereas the full softmax baseline needs $S=40$.
Aggressively splitting the mini-batch (i.e. using larger values of $S$) slows down the training time, and in that sense our method is much more efficient when we consider the larger mini-batch sizes.
If we increase the mini-batch size to 4096, our small softmax method works with $S=12$.

\section{Related Work}

Reducing the computational cost at the large softmax layer in language modeling/generation is actively studied~\citep{large_vocab,blackout,eriguchi2016,word_pred_arxiv,nce_gpu,word_pred_aaai}.
Most of the existing methods try to reduce the vocabulary size by either negative sampling or vocabulary prediction.
One exception is that \citet{oda2017} propose to predict a binary code of its corresponding target word.
Although such a sophisticated method is promising, we focused on the vocabulary reduction method to apply policy-based reinforcement learning in a straightforward way.

As reported in this paper, one simple way to define a reward function for reinforcement learning is to use task-specific automatic evaluation metrics~\citep{rnn_rl2016,gnmt,self_critic2017,simp_rl2017,romain2018}, but this is limited in that we can only use training data with gold target sentences.
An alternative approach is to use a {\it discriminator} in generative adversarial networks~\citep{gan}, and \citet{gan_rl2018} showed that REINFORCE with such a discriminator improves translation accuracy.
However, \citet{gan_rl2018} only used the training data, and thus the potential of the generative adversarial networks is not fully realized.
One promising direction is to improve the use of the generative adversarial networks for the sentence generation tasks by using our method, because our method can also accelerate the combination of REINFORCE and the discriminator.

\section{Conclusion}
This paper has presented how to accelerate reinforcement learning for sentence generation tasks by reducing large action spaces.
Our method is as accurate as, is faster than, and uses less GPU memory than the standard full softmax counterpart, on sentence generation tasks of different modalities.
In future work, it is interesting to use our method in generative adversarial networks to further improve the sentence generation models.

\section*{Acknowledgments}
We thank anonymous reviewers for their fruitful comments.
This work was supported by JST CREST Grant Number JPMJCR1513, Japan.


\bibliography{bibtex}
\bibliographystyle{acl_natbib}

\appendix

\section*{Supplementary Material}

\section{Vocabulary Prediction Model}

\paragraph{Residual block}
In Section~3.2, we used a residual block $r(X) = \mathrm{Res}(v(X))\in\mathbb{R}^{d_v}$ inspired by \citet{resnet} to transform the input vector $v(X)\in\mathbb{R}^{d_v}$:
\begin{equation}
\begin{split}
r_1 &= \mathrm{BN}_{r_1}(v(X)),~r_2 = \mathrm{tanh}(r_1),\\
r_3 &= W_{r_3} r_2 + b_{r_3},~~r_4 = \mathrm{BN}_{r_4}(r_3),\\
r_5 &= \mathrm{tanh}(r_4),~~~~~~~~r_6 = W_{r_6} r_5 + b_{r_6},\\
r(X) &= r_6 + v(X),
\end{split}
\end{equation}
where $\mathrm{BN}_{r_1}(\cdot)$ and $\mathrm{BN}_{r_4}(\cdot)$ correspond to batch normalization~\citep{bn}, $W_{r_3}\in\mathbb{R}^{d_v\times d_v}$ and $W_{r_6}\in\mathbb{R}^{d_v\times d_v}$ are weight matrices, and $b_{r_3}\in\mathbb{R}^{d_v}$ and $b_{r_6}\in\mathbb{R}^{d_v}$ are bias vectors.
We apply dropout~\citep{dropout} to $r_5$ with a dropout rate of 0.4.

\paragraph{Label smoothing}
In Section~3.2, we applied label smoothing~\citep{smoothing} to the loss function in Equation~(10).
More concretely, we modify the gold label $t_i$ for the $i$-th target word as follows:
\begin{eqnarray}
t_i \leftarrow (1.0-\varepsilon)t_i + \varepsilon p(i),
\end{eqnarray}
where $\varepsilon$ is a hyperparameter, and $p(i)$ is a prior probability that the $i$-th word appears in a target sentence.
$p(i)$ is computed for each dataset:
\begin{eqnarray}
p(i)=\frac{\sum_{j=1}^{|T|} t_i^j}{|T|},
\end{eqnarray}
where $|T|$ is the size of the training dataset, and $t_i^j$ is the gold label for the $i$-th target word in the $j$-th training example.
Therefore, $p(i)$ roughly reflects the unigram frequency.
We have empirically found that the recommended value $\varepsilon=0.1$ consistently improves the recall of the predictor.

\section{Detailed Experimental Settings}

\begin{table*}[t]
  \begin{center}
{\small
    \begin{tabular}{lcc|cc|cc}
					& & & \multicolumn{2}{c|}{CPU} & \multicolumn{2}{c}{GPU}  \\
					Data size & $|V|$ &  Model size & Small softmax & Full softmax & Small softmax & Full softmax \\ \hline
	100K			& 23,536 & 1-L, 256-D & ~~54.4  & 113.8 &  71.9 & ~~78.4   \\
	2M 				& 70,668 & 2-L, 512-D & 156.2 & 503.2 & 80.5 & 105.7 \\
	2M, SW		& 37,905 & 2-L, 512-D & 161.0 & 369.2 & 84.8 & ~~99.2  \\ \hline
    \end{tabular}
}
    \caption{Average time [milliseconds] to obtain a translation for each sentence in the En-Ja development split.}
    \label{tb:test_time}
  \end{center}
\end{table*}

\paragraph{Word segmentation}
The sentences in the En-Vi, MS COCO, and Flickr8K datasets were pre-tokenized.
We used the {\tt Kytea} toolkit for Japanese and the {\tt Stanford Core NLP} toolkit for Chinese.
In the other cases, we used the {\tt Moses} word tokenizer.
We lowercased all the English sentences.
The En-Ja (2M, SW) dataset was obtained by the {\tt SentencePiece} toolkit so that the vocabulary size becomes around 32,000.

\paragraph{Vocabulary construction}
We built the target language vocabularies with words appearing at least five times for the En-De dataset, seven times for the En-Ja (2M) dataset, three times for the Ch-Ja dataset, and twice for the other datasets.

\paragraph{Optimization}
We initialized all the weight and embedding matrices with uniform random values in $[-0.1, +0.1]$, and all the bias vectors with zeros, except for the LSTM forget-gate biases which were initialized with ones~\citep{bias1}.
For all the models, we used gradient-norm clipping~\citep{clip} with a clipping value of $1.0$.
We applied dropout to Equation~(3), (4), and (5) with a dropout rate of $0.2$, and we further used dropout in the vertical connections of the two-layer LSTMs~\citep{rnndrop} for the En-Ja (2M) and (2M, SW) datasets.
As regularization, we also used weight decay with a coefficient of $10^{-6}$.
When training the vocabulary predictor and the sentence generation models, we checked the corresponding evaluation scores at every half epoch, and halved the learning rate if the evaluation scores were not improved.
We stabilized the training of the sentence generation models by not decreasing the learning rate in the first six epochs.
These training settings were tuned for the En-Ja (100K) dataset, but we empirically found that the same settings lead to the consistent results for all the datasets.

\paragraph{Baseline Estimator}
We used the {\tt Adam} optimizer with a learning rate of $10^{-3}$ and the other default settings, to optimize the baseline estimator in Section~2.2.
We have found that our results are not sensitive to the training settings of the baseline estimator.

\paragraph{Beam search}
For the results in Table~3 and 4, we tried two beam search methods in \citet{lgp_nmt} and \citet{wat_oda}, and selected better scores for each setting.
In general, these length normalization methods lead to the best BLEU scores with a beam size of 10 to 20.

\section{Test Time Efficiency}

By the fact that our method works efficiently with reinforcement learning, we expect that our method also works well at test time.
Table~\ref{tb:test_time} shows the average decoding time [milliseconds] to generate a Japanese sentence given an English sentence for the En-Ja development split.
For reference, the vocabulary size and the model size are also shown for each setting.
We note that the decoding time of our method {\it includes} the time for constructing an input-specific vocabulary for each source input.

We can see that our method runs faster than ``Full softmax''; in particular, the speedup is significant on CPUs, and the decoding time by our method is less sensitive to changing $|V|$ than that of ``Full softmax''.
This is because our method handles the full vocabulary only once for each source input, whereas ``Full softmax'' needs to handle the full vocabulary every time the model predicts a target word.

\end{document}